# USING INTELLIGENT AGENTS TO UNDERSTAND MANAGEMENT PRACTICES AND RETAIL PRODUCTIVITY

Peer-Olaf Siebers
Uwe Aickelin

School of Computer Science, Automated Scheduling,
Optimisation and Planning Research Group (ASAP)
University of Nottingham
Nottingham, NG8 1BB, UK

Helen Celia
Chris W. Clegg

Centre for Organisational Strategy, Learning & Change,
Leeds University Business School
University of Leeds
Leeds, LS2 9JT, UK

**ABSTRACT**

Intelligent agents offer a new and exciting way of understanding the world of work. In this paper we apply agent-based modeling and simulation to investigate a set of problems in a retail context. Specifically, we are working to understand the relationship between human resource management practices and retail productivity. Despite the fact we are working within a relatively novel and complex domain, it is clear that intelligent agents could offer potential for fostering sustainable organizational capabilities in the future. The project is still at an early stage. So far we have conducted a case study in a UK department store to collect data and capture impressions about operations and actors within departments. Furthermore, based on our case study we have built and tested our first version of a retail branch simulator which we will present in this paper.

## 1 INTRODUCTION

The retail sector has been identified as one of the biggest contributors to the productivity gap that persists between the UK, Europe and the USA (Reynolds et al. 2005). It is well documented that measures of UK retail productivity rank lower than those of countries with comparably developed economies. Intuitively, it is inevitable that management practices are inextricably linked to a company's productivity and performance. However, many researchers have struggled to provide clear empirical evidence using more traditional research methods (for a review, see Wall and Wood 2005).

 Significant research has been done to investigate the productivity gap and the common focus has been to quantify its size and determine the contributing factors. Best practice guidelines have been developed and published, but there remains considerable inconsistency and uncertainty regarding how these are implemented and manifested in the retail work place. Siebers et al. (submitted) have conducted a comprehensive literature review of this pertinent research area linking management practices to firm-level productivity. Practices are dichotomized according to their focus, whether operationally-focused or people-focused. The authors conclude that management practices are multidimensional constructs that generally do not demonstrate a straightforward relationship with productivity variables. Empirical evidence affirms that management practices must be context specific to be effective, and in turn productivity indices must also reflect a particular organization's activities.

 Currently there is no reliable and valid way to delineate the effects of management practices from other socially embedded factors. Most Operational Research (OR) methods can be applied as analytical tools once management practices have been implemented, however they are not very useful at revealing system-level effects of the introduction of specific management practices. This holds particularly when the focal interest is the development of the system over time, like in the real world. This contrasts with more traditional techniques, which allow us to identify the state of the system at a certain point in time.

 The overall aim of our project is to understand and predict the impact of different management practices on retail store productivity. To achieve this aim we have adopted a case study approach and integrated applied research methods to collect both qualitative and quantitative data. In summary, we have conducted four weeks of informal participant observations, forty staff interviews supplemented by a short questionnaire on the effectiveness of various management practices, and drawn upon a variety of established informational sources internal to the case study organization. Using this data, we are applying Agent-Based Modeling and Simulation (ABMS) to try to devise a functional representation of the case study departments.

In this paper we will focus on the simulation side of the project. In Section 2 we summarize the literature review we have conducted to find a suitable research tool for our study. Section 3 describes the conceptualization, design and implementation of our retail branch simulator. In Section 4 we describe two experiments that we have conducted as a first step to validate our retail branch simulator. Section 5 concludes the paper and unveils our future ideas.

## 2 WHY AGENT-BASED SIMULATION?

OR is applied to problems concerning the conduct and coordination of the operations within an organization (Hillier and Lieberman 2005). An OR study usually involves the development of a scientific model that attempts to abstract the essence of the real problem. When investigating the behavior of complex systems the choice of an appropriate modeling technique is very important. In order to be able to make a choice for our project, we reviewed the relevant literature spanning the fields of Economics, Social Science, Psychology, Retail, Marketing, OR, Artificial Intelligence, and Computer Science. Within these fields a wide variety of approaches are used which can be classified into three main categories: analytical approaches, heuristic approaches, and simulation. In many cases we found that combinations of these were used within a single model. Common combinations were 'simulation / analytical' for comparing efficiency of different non-existing scenarios, (e.g. Greasley 2005), and 'simulation / analytical' or 'simulation / heuristic' where analytical or heuristic models were used to represent the behavior of the entities within the simulation model (e.g. Schwaiger and Stahmer 2003).

In our review we put a particular emphasis on those publications that try to model the link between management practices and productivity in the retail sector. We found a very limited number of papers that investigate management practices in retail at firm level. The majority of these papers focus on marketing practices (e.g. Keh et al. 2006). By far the most frequently used modeling technique we found being used was agent-based modeling employing simulation as the method of execution. It seems to be the natural way of system representation for these purposes.

Simulation introduces the possibility of a new way of thinking about social and economic processes, based on ideas about the emergence of complex behavior from relatively simple activities (Simon 1996). While analytical models typically aim to explain correlations between variables measured at one single point in time, simulation models are concerned with the development of a system over time. Furthermore, analytical models usually work on a much higher level of abstraction than simulation models. For simulation models it is critical to define the right level of abstraction. Csik (2003) states that on the one hand the number of free parameters should be kept on a level as low as possible. On the other hand, too much abstraction and simplification might threaten the homomorphism between reality and the scope of the simulation model. There are several different approaches to simulation, amongst them discrete event simulation, system dynamics, micro simulation and agent-based simulation. The choice of the most suitable approach always depends on the issues investigated, the input data available, the level of analysis and the type of answers to be sought.

Although computer simulation has been used widely since the 1960s, ABMS only became popular in the early 1990s (Epstein and Axtell 1996). ABMS can be used to study how micro-level processes affect macro-level outcomes. A complex system is represented by a collection of individual agents that are programmed to follow simple behavioral rules. Agents can interact with each other and with their environment to produce complex collective behavioral patterns. Macro behavior is not explicitly simulated; it emerges from the micro-decisions made by the individual agents (Pourdehnad et al. 2002). The main characteristics of agents are their autonomy, their ability to take flexible action in reaction to their environment and their pro-activeness depending on motivations generated from their internal states. They are designed to mimic the attributes and behaviors of their real-world counterparts. The simulation output may be potentially used for explanatory, exploratory and predictive purposes (Twomey and Cadman 2002). This approach offers a new opportunity to realistically and validly model organizational characters and their interactions, to allow a meaningful investigation of human resource management practices. ABMS is still a relatively new simulation technology and its principle application has been in academic research. With the appearance of more sophisticated modeling tools in the broader market, things are starting to change (Luck et al. 2005). Also, an ever increasing number of computer games use the ABMS approach.

A detailed description of ABMS and thoughts on the appropriate contexts for ABMS versus conventional modeling techniques can be found in WSC introductory tutorial on ABMS (Macal and North 2006). Therefore we provide only a brief summary of some of our thoughts.

Due to the characteristics of the agents, this modeling approach appears to be more suitable than Discrete Event Simulation (DES) for modeling human-oriented systems (Siebers 2006). ABMS seems to promote a natural form of modeling, as active entities in the live environment are interpreted as actors in the model. There is a structural correspondence between the real system and the model representation, which makes them more intuitive and easier to understand than for example a system of differential equations as used in System Dynamics. Hood (1998) emphasized that one of the key strengths of ABMS is that the system as a whole is not constrained to exhibit any particular behavior as the system properties emerge from its constituent agent interactions. Consequently assumptions of linearity, equilibrium and so on, are not needed. With regard to disadvantages there is a general consensus in the literature that it is difficult to evaluate agent-based models, because the behavior of the system emerges from the interactions between the individual entities. Furthermore, problems often occur through the lack of adequate empirical data. Finally, there is always the danger that people new to ABMS may expect too much from the models, particularly in regard to predictive ability.

## 3 MODEL DESIGN AND IMPLEMENTATION

The strategy for our project is iterative, creating a relatively simple model and then building in more and more complexity. To begin with we have been trying to understand the particular problem domain, to generate the underlying rules currently in place. We are now in the process of building an agent based simulation model of the real system using the information gathered during our case study and will then validate our model by simulating the operation of the real system. This approach will allow us to assess the accuracy of the system representation. If the simulation provides a sufficiently good representation we are able to move to the next stage, and generate new scenarios for how the system could work using new rules.

### 3.1 Modelling Concepts

Our case study approach and analysis has played a crucial role allowing us to acquire a conceptual idea of how the real system is structured. This is an important stage of the project, revealing insights into the operation of the system as well as the behavior of and interactions between the different characters in the system. We have designed the system by applying a DES approach to conceptualize and model the system, and then an agent approach to conceptualize and model the actors within the system. This method made it easier to design the model, and is possible because only the actors' action requires an agent based approach.

In terms of performance indicators, these are identical to those of a DES model. Beyond this, ABMS can offer further insights. A simulation model can detect unintended consequences, which have been referred to as 'emergent behavior' (Gilbert and Troitzsch 2005). Such unintended consequences can be difficult to understand because they are not defined in the same way as the system inputs; however it is critical to fully understand all system output to be able to accurately draw comparisons between the relative efficiencies of competing systems.

Our conceptual ideas for the simulator are shown in Figure 1. Within our simulation model we have three different types of agents (customers, sales staff, and managers) each of them having a different set of relevant parameters. We will use probabilities and frequency distributions to assign slightly different values to each individual agent. In this way a population is created that reflects the variations in attitudes and behaviors of their real human counterparts. In terms of other inputs, we need global parameters which can influence any aspect of the system, and may for example define the number of agents in the system. With regards to the outputs we always hope to find some unforeseeable, emergent behavior on a macro level. Having a visual representation of the simulated system and its actors will allow us to monitor and better understand the interactions of entities within the system. Coupled with the standard DES performance measures, we hope to identify bottlenecks and help to optimize the modeled system.

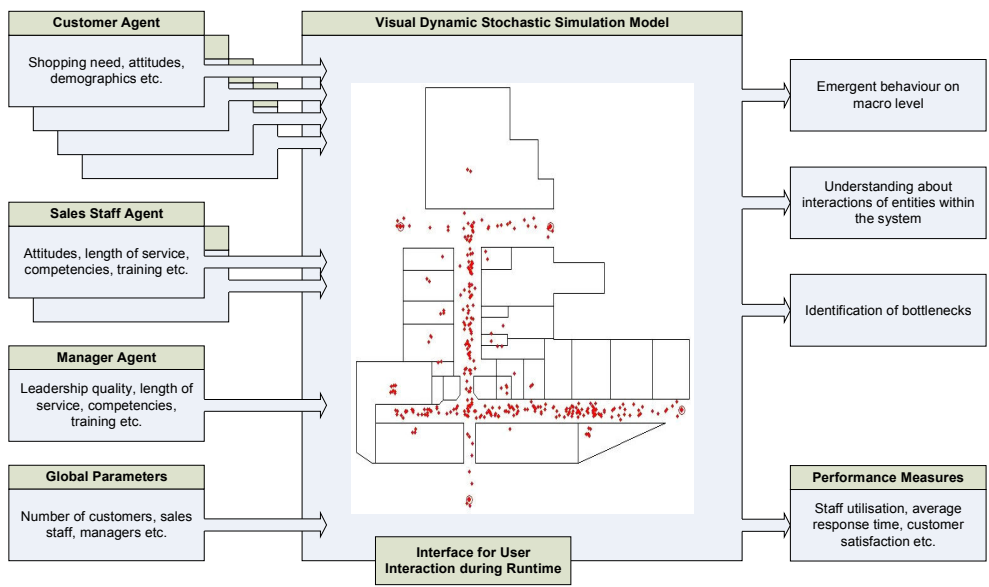

Figure 1: Conceptual model for our simulator

For the conceptual design of our agents we have decided to use state charts. State charts show the different states an entity can be in and also define the events that cause a transition from one state to another. This is exactly the information we need in order to represent our agents later within the simulation environment. Furthermore, this form of graphical representation is also helpful for validating the agent design as it is easier for non-specialists to understand.

The art of modelling pivots on simplification and abstraction (Shannon 1975). A model is always a restricted copy of the real world, and we have to identify the most important components of a system to build effective models. In our case, instead of looking for components we have to identify the most important behaviours of an actor and the triggers that initiate a move from one state to another. We have developed state charts for all the relevant actors in our retail branch model. Figure 2 shows as an example the state charts for a customer agent. The transition rules have been replaced by numbers to keep the chart comprehensible. They are explained in detail in the Section 3.2.

A question that can be asked is whether our agents are intelligent or not? Wooldridge (2002) states that in order to be intelligent agents need to have the following attributes: being reactive, being proactive and being social. This is a widely accepted view. Being reactive means responding to changes in the environment (in a timely manner), while being proactive means persistently pursuing goals and being social means interacting with other agents (Padgham and Winikoff, 2004). Our agents perceive a goal in that they want to either buy something or return something. For buying they have a sub goal; that they are trying to buy the right thing. If they are not sure they will ask for help. Our agents are not only reactive but also flexible, i.e. they are capable to recover from a failure of action. They have alternatives inbuilt when they are unable to perceive their goal, e.g. if they want to pay and things are not moving forward in the queue they always have the chance to leave a queue and continue with another action. They are responding in a flexible way to certain changes in their environment, in this case the length of the queue. Finally, as there is communication between agents and staff, they can also be regarded as being social.

### 3.2 Empirical Data

Often agents are based on analytical models or heuristics and in the absences of adequate empirical data theoretical models are employed. However, for our agents we use frequency distributions for state change delays and probability distributions for decision making processes as statistical distributions are the best format to represent the data we have gathered during our case study due to their numerical nature. The case study was conducted in the Audio and Television (A&TV) and the WomensWear (WW) departments of a leading UK department store. As mentioned earlier we have conducted informal participant observations, staff interviews, and drawn upon a variety of established informational sources internal to the case study organization.

Our frequency distributions are modeled as triangular distributions supplying the time that an event lasts, using the minimum, mode, and maximum duration. Our triangular distributions are based on our own observation and expert estimates in the absence of numerical data. We have collected this information from the two branches and calculated an average value for each department type, creating one set of data for A&TV and one set for WW. Table 1 lists some sample frequency distributions that we have used for modeling the A&TV department (the values presented here are slightly amended to comply with con-fidentiality restrictions). The distributions are used as exit rules for most of the states. All remaining exit rules are based on queue development, i.e. the availability of staff.

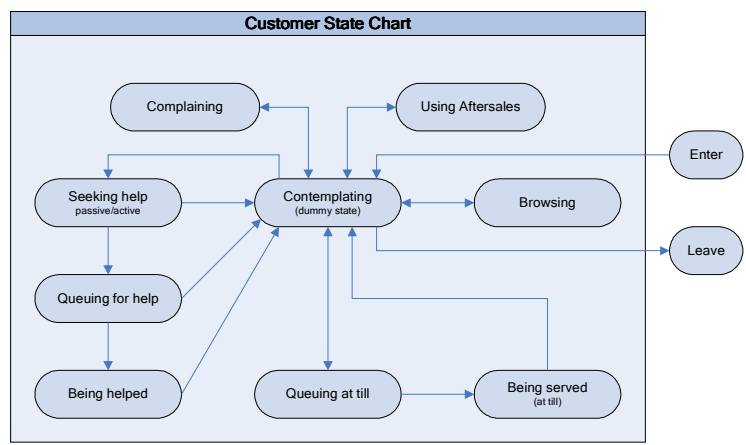

Figure 2: Conceptual model for customer agent

| situation | min | mode | max |
|---|---|---|---|
| leave browse state after … | 1 | 7 | 15 |
| leave help state after … | 3 | 15 | 30 |
| leave pay queue (no patience) after … | 5 | 12 | 20 |

Table 1: Sample frequency distribution values

The probability distributions are partly based on company data (e.g. conversion rates, i.e. the percentage of customers who buy something) and partly on informed guesses (e.g. patience of customers before they would leave a queue). As before, we have calculated average values for each department type. Some examples for probability distributions we used to model the A&TV department can be found in Table 2. The distributions make up most of the transition rules at the branches where decisions are made with what action to perceive (e.g. decision to seek help). The remaining decisions are based on the state of the environment (e.g. leaving the queue, if the queue does not get shorter quickly enough).

| event | probability it occurs |
|---|---|
| someone makes a purchase after browsing | 0.37 |
| someone requires help | 0.38 |
| someone makes a purchase after getting help | 0.56 |

Table 2 – Sample probabilities

Company data is available about work team numbers and work team composition, varying opening hours and peak times (to be implemented in future). Also financial data (e.g. transaction numbers and values) are available but have not been used at this stage.

### 3.3 Implementation

Our simulation has been implemented in AnyLogic™ which is a Java™ based multi-paradigm simulation software (XJ Technologies 2007). During the implementation we have used the knowledge, experience and data gained through our case study work. The simulator can represent the following actors: customers, service staff (including cashiers, selling staff of two different training levels) and managers. Figure 3 shows a screenshot of the current customer and staff agent logic as it has been implemented in AnyLogic™. Boxes show customer states, arrows possible transitions and numbers satisfaction weights.

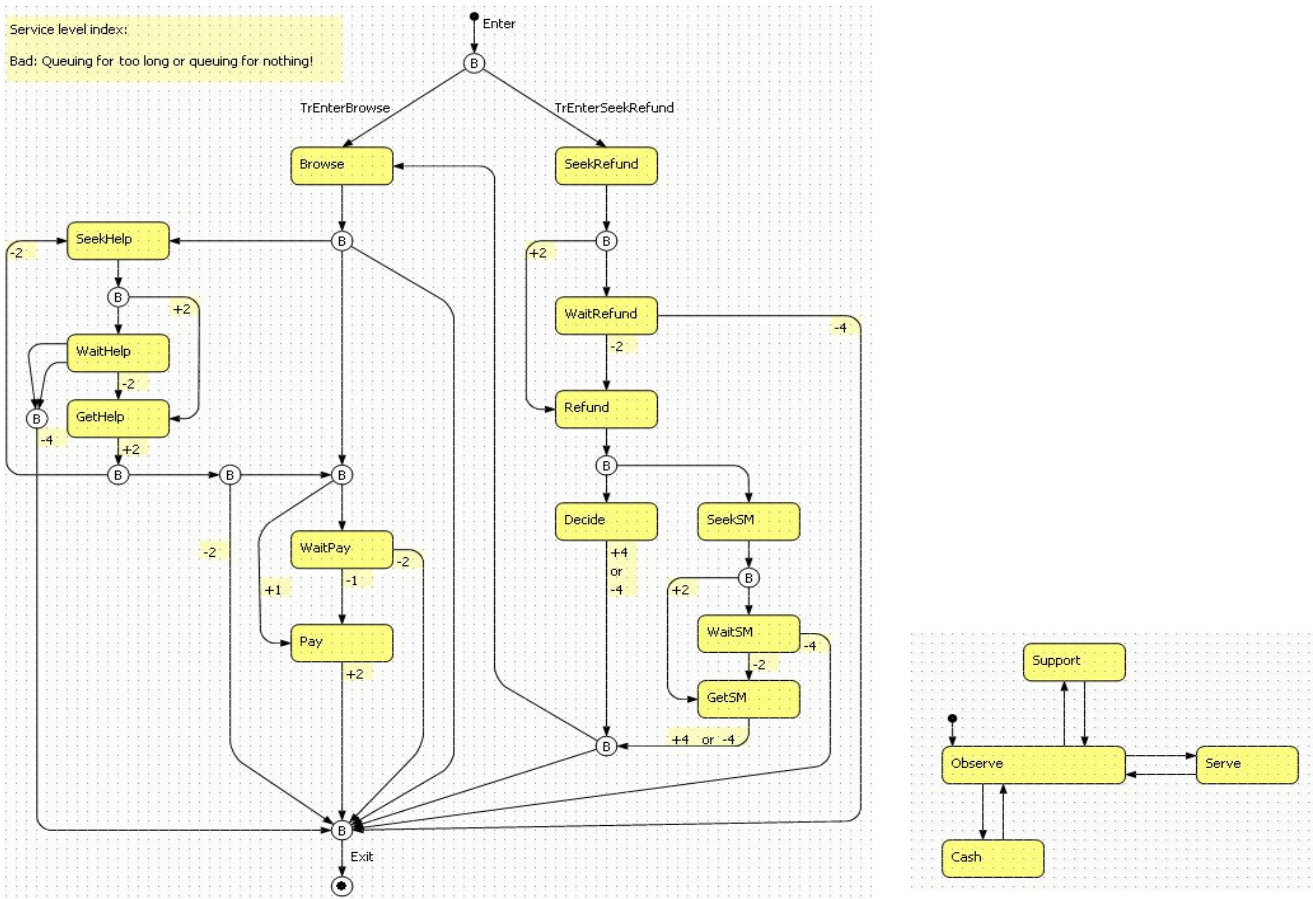

Figure 3: Customer (left) and staff (right) agent logic implementation in AnyLogic™

Currently there are two different types of customer goals implemented: making a purchase or obtaining a refund. If a refund is granted, the customer's goal may then change to making a new purchase, or alternatively they will leave the shop straight away. The customer agent template consists of three main blocks which all use a very similar logic. These blocks are 'Help', 'Pay' and 'Refund'. In each block, in the first instance, customers will try to obtain service directly and if they cannot obtain it (no suitable staff member available) they will have to queue. They will then either be served as soon as the right staff member becomes available or they will leave the queue if they do not wait any longer (an autonomous decision). A complex queuing system has been implemented to support different queuing rules. In comparison to the customer agent template, the staff agent template is relatively simple. Whenever a customer requests a service and the staff member is available and has the right level of expertise for the task requested, the staff member commences this activity until the customer releases the staff member. While the customer is the active component of the simulation model the staff member is currently passive, simply reacting to requests from the customer. In future we planned to add a more pro-active role for the staff members, e.g. offering services to browsing customers.

A service level index is introduced as a new performance measure. The index allows customer service satisfaction to be recorded throughout the simulated lifetime. The idea is that certain situations might have a bigger impact on customer satisfaction than others, and therefore differential weightings are assigned to events to account for this. For example, in our model if a customer starts to wait for a refund and leaves without one, then their satisfaction index decreases by 4 (see figure 3). We measure customer satisfaction in two different ways derived from these weightings; both in terms of how many customers leave the store with a positive service level index value, and the sum of all customers' service level index values. Applied in conjunction with an ABMS approach, we expect to observe interactions with individual customer differences, variations which have been empirically linked to differences in customer satisfaction. This helps the analyst to find out to what extent customers underwent a positive or negative shopping experience. It also allows the analyst to put emphasis on different operational aspects and try out the impact of different strategies.

The simulator can be initialized from an Excel™ spreadsheet and supports the simulation of the two types of departments we looked at during our case study. These differ with respect to their staffing, service provision and customer requirements, which we hope will be reflected in the simulation results. WW customers will ask for help when they know what they want whereas A&TV customers will ask for help when they do not know what they want. WW makes a lot more unassisted sales than A&TV and service times are very different; in WW the average service time is a lot shorter than in A&TV. This service requirement has a differential impact on the profile of employee skills at the department level.

## 4 A FIRST VALIDATION OF OUR SIMULATOR

To test the operation of our simulator and ascertain face validity we have designed and run 2 sets of experiments for both departments. Our case study work has helped us to identify the distinguishing characteristics of the departments, for example different customer arrival rates and different service times. In these experiments we will examine the impact of these individual characteristics on the volume of sales transactions and customer satisfaction indices. All experiments hold the overall number of staffing resources constant at 10 staff and we run the simulation for a period of 10 weeks. We have conducted 20 repetitions for every experimental condition enabling the application of rigorous statistical techniques.

Each set of results are analyzed for each dependent variable using a two-way between-groups analysis of variance (ANOVA). Despite our prior knowledge of how the real system operates, we were unable to hypothesize precise differences in variable relationships, instead predicting general patterns of relationships. Indeed, ABMS is a decision-support tool and is only able to inform us about directional changes between variables (actual figures are notional). Where significant ANOVA results were found, post-hoc tests were applied where possible to investigate further the precise impact on outcome variables under different experimental conditions.

During our time in the case study organization, we observed that over time the number of cashiers available to serve customers would fluctuate. In the first experiments we vary the staffing arrangement (i.e. the number of cashiers) and examine the impact on the volume of sales transactions and two levels of customer satisfaction; both customer satisfaction (how many customers leave the store with a positive service level index value) and overall satisfaction (the sum of all customers' service level index values). In reality, we saw that allocating extra cashiers would reduce the shop floor sales team numbers, and therefore the total number of customer-facing staff in each department is kept constant at 10. We therefore predict that for each of our dependent measures: number of sales transactions (1), customer satisfaction index (2) and overall satisfaction index (3):

- Ha: An increase in the number of cashiers will be linked to increases in 1, 2 and 3 to a peak level, beyond which 1, 2 and 3 will decrease.
- Hb: The peak level of 1, 2 and 3 will occur with a smaller number of cashiers in A&TV than in WW.

| Department | Cashiers | Number of Transactions | | Customer Satisfaction | | Overall Satisfaction | |
|---|---|---|---|---|---|---|---|
| | | mean | std. dev. | mean | std. dev. | mean | std. dev. |
| A&TV | 1 | 4853.50 | 26.38 | 12324.05 | 77.64 | 9366.40 | 563.88 |
| | 2 | 9822.20 | 57.89 | 14762.45 | 81.04 | 19985.20 | 538.30 |
| | 3 | 14279.90 | 96.34 | 17429.70 | 103.77 | 28994.80 | 552.60 |
| | 4 | 14630.60 | 86.19 | 17185.00 | 99.09 | 32573.60 | 702.64 |
| | 5 | 13771.85 | 97.06 | 16023.20 | 82.66 | 27916.05 | 574.56 |
| WW | 1 | 8133.75 | 22.16 | 18508.20 | 88.68 | 17327.95 | 556.03 |
| | 2 | 15810.10 | 56.16 | 22640.40 | 92.00 | 42339.10 | 736.61 |
| | 3 | 25439.60 | 113.66 | 28833.10 | 115.65 | 58601.10 | 629.68 |
| | 4 | 30300.70 | 249.30 | 32124.60 | 230.13 | 74233.30 | 570.79 |
| | 5 | 28894.25 | 195.75 | 30475.20 | 176.41 | 76838.65 | 744.31 |
| Total | 1 | 6493.63 | 1661.19 | 15416.13 | 3132.55 | 13347.18 | 4069.20 |
| | 2 | 12816.15 | 3032.61 | 18701.43 | 3990.07 | 31162.15 | 11337.24 |
| | 3 | 19859.75 | 5651.89 | 23131.40 | 5775.35 | 43797.95 | 15003.13 |
| | 4 | 22465.65 | 7937.00 | 24654.80 | 7566.98 | 53403.45 | 21104.67 |
| | 5 | 21333.05 | 7659.04 | 23249.20 | 7319.32 | 52377.35 | 24781.61 |

Table 3: Descriptives for first experiments (all to 2 d.p.)

An ANOVA was run for each dependent variable, and all revealed statistically significant differences (see Table 3 for descriptive statistics). For 1 and 2, Levene's test for equality of variances was violated ($p<.05$) so a more stringent significance level was set ($p<.01$).

For 1 there were significant main effects for both department [$F(1, 190) = 356441.1$, $p<.001$] and staffing [$F(4, 190) = 124919.5$, $p<.001$], plus a significant interaction effect [$F(4, 190) = 20496.37$, $p<.001$]. Tukey's post hoc tests for the impact of staffing revealed significant differences for every single comparison ($p<.001$).

There is clear support for H1a. We expected this to happen because the number of normal staff available to provide customer advice will eventually reduce to the extent where there will be a detrimental impact on the number of customers making a purchase. Some customers will become impatient waiting increasingly long for service, and will leave the department without making a purchase. H1b is not supported, the data presents an interesting contrast, in that 1 plateaus in A&TV around 3 and 4 cashiers, whereas WW benefits greatly from the introduction of a fourth cashier. Nonetheless this finding supports the thinking underlying this hypothesis, in that we expected the longer average service times in A&TV to put a greater 'squeeze' on customer advice with even a relatively small increase in the number of cashiers.

For 2, there were significant main effects for both department [$F(1, 190) = 391333.7$, $p<.001$], and staffing [$F(4, 190) = 38633.83$, $p<.001$], plus a significant interaction effect [$F(4, 190) = 9840.07$, $p<.001$]. Post hoc tests for staffing revealed significant differences for every single comparison ($p<.001$).

The results support both H2a and H2b. We interpret these findings in terms of A&TV's greater service requirement, combined with the reduced availability of advisory sales staff. These factors result in a peak in purchasing customers' satisfaction with a smaller number of cashiers (4) than in WW (5).

For 3, there were significant main effects for both department [$F(1, 190) = 117214.4$, $p<.001$], and staffing [$F(4, 190) = 29205.09$, $p<.001$], plus a significant interaction effect [$F(4, 190) = 6715.93$, $p<.001$]. Tukey's post hoc comparisons indicated significant differences between all staffing levels ($p<.001$).

Our results support H3a for A&TV, showing a clear peak in overall satisfaction. H3a is only partially supported for WW, in that no decline in 3 is evident with up to 5 cashiers, although increasing this figure may well expose a peak because the overall satisfaction appears to be starting to plateau out. The results offer firm support in favor of H3b.

The second experiment investigates employee empowerment. During our case study we observed the implementation of a new refund policy. This new policy allows cashiers to independently decide whether or not to make a refund up to the value of £50, rather than referring the authorization decision to a section manager. To simulate this practice, we vary the probability that cashiers are empowered to make refund decisions autonomously. We assess its impact in terms of two performance measures: overall customer refund satisfaction and cashier utilization (a proportion of maximum capacity). The staffing arrangement is held constant, consisting of 3 cashiers, 5 normal staff members, 1 expert staff member, and 1 section manager. We hypothesize that:

- H4. Higher levels of empowerment will be linked to higher refund satisfaction.
- H5. Higher levels of empowerment will be linked to greater cashier utilization.

An ANOVA was run for each outcome measure (see Table 4 for descriptives). For refund satisfaction, there were significant main effects for both department [$F(1, 190) = 508.73$, $p<.001$], and empowerment [$F(4, 190) = 120.46$, $p<.001$], plus a significant interaction effect [$F(4, 190) = 29.81$, $p<.001$]. Tukey's post hoc tests for the impact of empowerment revealed significant differences between all comparisons ($p<.001$), except for .00 with .75, and .25 with .50, where there were no significant differences.

| Department | Empower-ment | Overall refund | | Cashier utilization | |
|---|---|---|---|---|---|
| | | mean | std. dev. | mean | std. dev. |
| A&TV | 0.00 | 3130.70 | 242.58 | 0.6286 | 0.00 |
| | 0.25 | 3880.70 | 225.70 | 0.6392 | 0.00 |
| | 0.50 | 3876.50 | 181.47 | 0.6488 | 0.00 |
| | 0.75 | 3716.80 | 225.31 | 0.6571 | 0.00 |
| | 1.00 | 2991.60 | 245.15 | 0.6623 | 0.00 |
| WW | 0.00 | 3090.30 | 222.00 | 0.6756 | 0.00 |
| | 0.25 | 3116.00 | 266.70 | 0.6737 | 0.00 |
| | 0.50 | 3041.20 | 211.75 | 0.6736 | 0.00 |
| | 0.75 | 2716.80 | 217.79 | 0.6722 | 0.00 |
| | 1.00 | 2085.20 | 168.19 | 0.6720 | 0.00 |
| Total | 0.00 | 3110.50 | 230.43 | 0.6521 | 0.02 |
| | 0.25 | 3498.35 | 457.61 | 0.6565 | 0.02 |
| | 0.50 | 3458.85 | 465.61 | 0.6612 | 0.01 |
| | 0.75 | 3216.80 | 551.59 | 0.6646 | 0.01 |
| | 1.00 | 2538.40 | 503.70 | 0.6672 | 0.01 |

Table 4: Descriptives for the third experiment (all to 2 d.p., except cashier utilization to 4 d.p.)

The data provides support for H4 between .00 and .25 levels of empowerment. However, as empowerment increases all of the results do not support our hypothesis, and demonstrate a counterintuitive progressive decline of refund satisfaction beyond the .25 level. Both departments display the curvilinear relationship between these two variables; refund satisfaction peaks at a middling level of empowerment (.50 for A&TV, .25 for WW). These results suggest that some constraining factors are occurring at the higher levels of empowerment. This may be linked to the empowered employees adhering to a stricter refund policy (resulting in less customer satisfaction), or the empowered employees taking longer to process the transaction.

For cashier utilization, there were significant main effects for both department $[F(1, 190) = 2913.45, p<.001]$, and empowerment $[F(4, 190) = 126.37, p<.001]$, plus a significant interaction effect $[F(4, 190) = 190.64, p<.001]$. Tukey's post hoc tests for the impact of cashier utilization confirmed significant differences between all comparisons ($p=.01$ between .75 and 1.00, $p<.001$ for all others).

The results support H5 for A&TV, but not for WW. In WW, empowerment is significantly inversely related to till utilization. Case study observations indicated that A&TV cashiers, like A&TV sales staff, when they have a higher level of empowerment they are motivated to work more efficiently. Empirical evidence indicated that WW cashiers may not be under the same time pressures to work more quickly, however this data goes one step further and suggests an inverse relationship.

## 5 CONCLUSIONS AND FUTURE DIRECTIONS

In this paper we present the conceptual design, implementation and operation of a retail branch simulator used to understand the impact of management practices on retail productivity. As far as we are aware this is the first time researchers have tried to use agent-based approaches to simulate management practices such as training and empowerment. Although our simulator uses specific case studies as source of information, we believe that the general model could be adapted to other retail companies and areas of management practices that have a lot of human interaction.

From what we can conclude from our current analyses, some findings are as hypothesized whereas others are more mixed. Further experimentation is required to enable rigorous statistical evolution of the outcome data and identification of statistically significant differences.

Currently we are developing our agents with the intention of enhancing their intelligence and heterogeneity. For this purpose we are introducing evolution and stereotypes. The most interesting system outcomes evolve over time and many of the goals of the retail company (e.g. service standards) are also planned long term. We are introducing an evolution of entities over time, including product knowledge for staff. Moreover, the customer population pool will be fixed to monitor customer agents over time. This allows us to consider shopping experience based on long-term satisfaction scores, with the overall effect being a certain 'reputation' for the shop. Another interesting aspect we are currently implementing is the introduction of stereotypes. Our case study organization has identified its particular customer stereotypes through market research. It will be interesting to find out how populations of certain customer types influence sales.

Overall, we believe that researchers should become more involved in this multi-disciplinary kind of work to gain new insights into the behavior of organizations. In our view, the main benefit from adopting this approach is the improved understanding of and debate about a problem domain. The very nature of the methods involved forces researchers to be explicit about the rules underlying behavior and to think in new ways about them. As a result, we have brought work psychology and agent-based modeling closer together to form a new and exciting research area.

**AUTHOR BIOGRAPHIES**

**PEER-OLAF SIEBERS** is a Research Fellow in the School of Computer Science and IT at the University of Nottingham. His main research interest is the application of computer simulation to study human oriented complex adaptive systems. Complementary fields of interest include distributed artificial intelligence, biologically inspired computing, game character behavior modeling, and agent-based robotics. His webpage can be found via `<www.cs.nott.ac.uk/~pos>`.

**UWE AICKELIN** is a Reader and Advanced EPSRC Research Fellow in the School of Computer Science and IT at the University of Nottingham. His research interests are mathematical modeling, agent-based simulation, heuristic optimization and artificial immune systems. see his webpage for more details `<www.aickelin.com>`.

**HELEN CELIA** is a Researcher at the Centre for Organisational Strategy, Learning and Change at Leeds University Business School. She is interested in developing ways of applying work psychology to better inform the modeling of complex systems using agents. For more information visit `<www.leeds.ac.uk/lubs/coslac>`

**CHRIS W. CLEGG** is a Professor of Organizational Psychology at Leeds University Business School. And the Deputy Director of the Centre for Organisational Strategy, Learning and Change. His research interests include: new technology, systems design, information and control systems, socio-technical thinking and practice; organizational change, change management, technological change; the use and effectiveness of modern management practices, innovation, productivity; new ways of working, job design, work organization. An emerging research interest is modeling and simulation. His webpage can be found via: `<www.leeds.ac.uk/lubs/coslac>`